\documentclass[sigconf, screen, nonacm]{acmart}

\AtBeginDocument{%
  }

\usepackage{soul}
\usepackage{multirow}
\usepackage{arydshln} 
\usepackage{makecell}
\usepackage{algorithm}
\usepackage{algorithmic}
\usepackage{array}

\definecolor{baselineRed}{HTML}{C00000}
\definecolor{incrementalGold}{HTML}{996600}

\begin{document}

\title{HyperAlign: Hyperbolic Entailment Cones for Adaptive Text-to-Image Alignment Assessment}


\author{Wenzhi Chen}
\email{d250201005@stu.cqupt.edu.cn}
\affiliation{%
  \institution{Chongqing University of Posts and Telecommunications}
  \city{Chongqing}
  \country{China}}

\author{Bo Hu}
\email{hubo90@cqupt.edu.cn}
\affiliation{%
  \institution{Chongqing University of Posts and Telecommunications}
  \city{Chongqing}
  \country{China}}

\author{Leida Li}
\email{ldli@xidian.edu.cn}
\affiliation{%
  \institution{Xidian University}
  \city{Xi'an}
  \country{China}}

\author{Lihuo He}
\email{lhhe@mail.xidian.edu.cn}
\affiliation{%
  \institution{Xidian University}
  \city{Xi'an}
  \country{China}}

\author{Wen Lu}
\email{luwen@mail.xidian.edu.cn}
\affiliation{%
  \institution{Xidian University}
  \city{Xi'an}
  \country{China}}

\author{Xinbo Gao}
\authornote{Corresponding author.}
\email{gaoxb@cqupt.edu.cn}
\affiliation{%
  \institution{Chongqing University of Posts and Telecommunications}
  \city{Chongqing}
  \country{China}}
\affiliation{%
  \institution{Xidian University}
  \city{Xi'an}
  \country{China}}

\renewcommand{\shortauthors}{Chen et al.}

\begin{abstract}
With the rapid development of text-to-image generation technology, accurately assessing the alignment between generated images and text prompts has become a critical challenge. Existing methods rely on Euclidean space metrics, neglecting the structured nature of semantic alignment and lacking adaptive capabilities for different samples, which may lead to misjudgments of fine-grained semantic relationships and unreliable alignment estimates across diverse generation scenarios. To address these limitations, we propose HyperAlign, an adaptive text-to-image alignment assessment framework based on hyperbolic entailment cones. First, we extract Euclidean features using CLIP and map them to hyperbolic space. Second, we design a dynamic-supervision entailment modeling mechanism that transforms discrete entailment logic into continuous geometric structure supervision. Finally, we propose an adaptive modulation regressor that utilizes hyperbolic geometric features to generate sample-level modulation parameters, adaptively calibrating Euclidean cosine similarity to predict the final score. Notably, HyperAlign introduces only lightweight components, maintaining low computational overhead. HyperAlign achieves highly competitive performance on both single-database evaluation tasks and cross-database generalization tasks, fully validating the effectiveness of hyperbolic geometric modeling for image-text alignment assessment.
\end{abstract}



\maketitle

\section{Introduction}

Recently, Text-to-Image (T2I) generation models such as Stable Diffusion~\cite{rombach2022ldm}, DALL-E~\cite{ramesh2021dalle}, and Midjourney have achieved leapfrog progress, driving production paradigm shifts in domains including media creation, artistic design, and advertising. However, the alignment degree between generated images and text prompts varies considerably. Unlike Natural Scene Image Quality Assessment (NS-IQA), users primarily focus on the degree of content alignment between generated images and input text prompts. Only when images faithfully reflect user intentions do considerations of aesthetic appeal and realism become meaningful. Therefore, establishing automated, high-precision Text-to-Image Alignment Assessment (T2IAA) methods not only provides alignment feedback to users but also serves as reward models to guide reinforcement learning optimization of generative models~\cite{hu2025towards}, holding significant research value.

Traditional NS-IQA methods, ranging from handcrafted methods~\cite{mittal2012brisque,mittal2013niqe} to deep learning methods~\cite{zhang2020blind}, have produced numerous classical methods over decades of development. Although the aforementioned methods perform well in assessing natural scene images, their performance on T2IAA tasks is lackluster due to their inability to model cross-modal semantic correlations. Existing methods for T2IAA mostly use multimodal large models like CLIP~\cite{radford2021clip} to map images and text into a unified representation space. These studies generally follow two paradigms: one is similarity-based metrics, such as AMFF-Net~\cite{zhou2024adaptive} and CIA-Net~\cite{zhou2025cia}, which directly convert cosine similarity to alignment scores; the other is learning-based regression-driven methods, such as IP-IQA~\cite{qu2024bringing}, which fine-tune feature extractors and train regression heads to predict scores. However, they still suffer from the following core limitations: (1) Existing methods compute feature similarities in flat Euclidean space, essentially treating alignment as simple symmetric distance metrics. This flattened representation ignores the structured nature of semantic alignment. (2) These methods typically employ fixed mapping functions, ignoring the substantial complexity and characteristic variations across different samples. This one-size-fits-all method leads to limited generalization when facing complex and diverse image-text pairs.

\begin{figure}[t]
\centering
\includegraphics[width=0.4\textwidth]{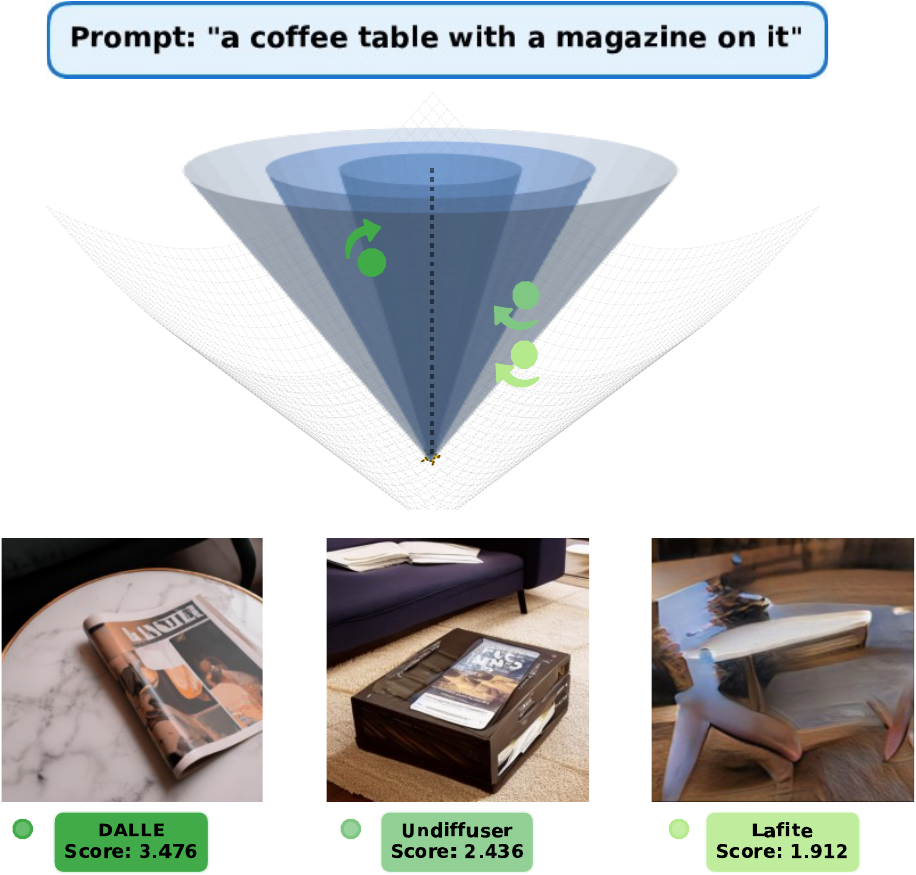}
\caption{Illustration of the proposed geometric constraint mechanism.
}
\Description{Illustration of hyperbolic entailment cones showing how high-alignment samples are constrained within narrower text cones while low-alignment samples have wider tolerance.}
\label{fig:intro}
\end{figure}

To address these issues, we propose HyperAlign, a T2IAA method based on hyperbolic entailment geometry. The core idea is to leverage the negative curvature property of hyperbolic space to explicitly model the hierarchical and asymmetric entailment relations in image-text pairs. For instance, ``Corgi'' entails ``dog'' entails ``animal'', but not vice versa, this is a strict partial order relationship that simple Euclidean similarity struggles to capture. Specifically, HyperAlign first projects CLIP features onto the Lorentz hyperbolic manifold, where image-text relations are decoupled into three interpretable geometric features: exterior angle, entailment aperture, and hyperbolic distance. Then, by introducing ground-truth alignment scores as geometric constraint signals, as illustrated in Figure~\ref{fig:intro}, we force high-alignment samples within narrower text cones while allowing low-alignment samples wider tolerance, transforming discrete entailment logic into continuous geometric supervision, thereby effectively modeling the hierarchical and asymmetric entailment structures. Finally, we design an adaptive modulation regressor that generates sample-level modulation parameters from hyperbolic geometric features to adaptively calibrate Euclidean cosine similarity for alignment score prediction. Notably, the additional components introduced by HyperAlign are intentionally kept lightweight: the hyperbolic mapping relies on closed-form algebraic operations, and the modulation regressor is a compact parameter prediction network, ensuring that the modeling gains do not come at the cost of heavy computational overhead.
The main contributions of this work can be summarized as follows:

\begin{itemize}

\item We propose the first T2IAA method that incorporates hyperbolic entailment cones with a dynamic-supervision entailment modeling mechanism. This method overcomes Euclidean-space limitations in modeling asymmetric entailment relations, transforms abstract score regression into intuitive geometric constraint learning, and achieves deep coupling between alignment degree and spatial structure.

\item We present an adaptive modulation regressor that utilizes entailment geometric features to generate sample-level modulation parameters, enabling personalized score mapping and improving generalization across complex and diverse samples.

\item Extensive experiments on multiple mainstream T2IAA benchmark datasets demonstrate that HyperAlign achieves highly competitive performance on both single-database evaluation and cross-database generalization tasks, while maintaining favorable computational efficiency.

\end{itemize}

\begin{figure*}[t]
\centering
\includegraphics[width=0.95\textwidth]{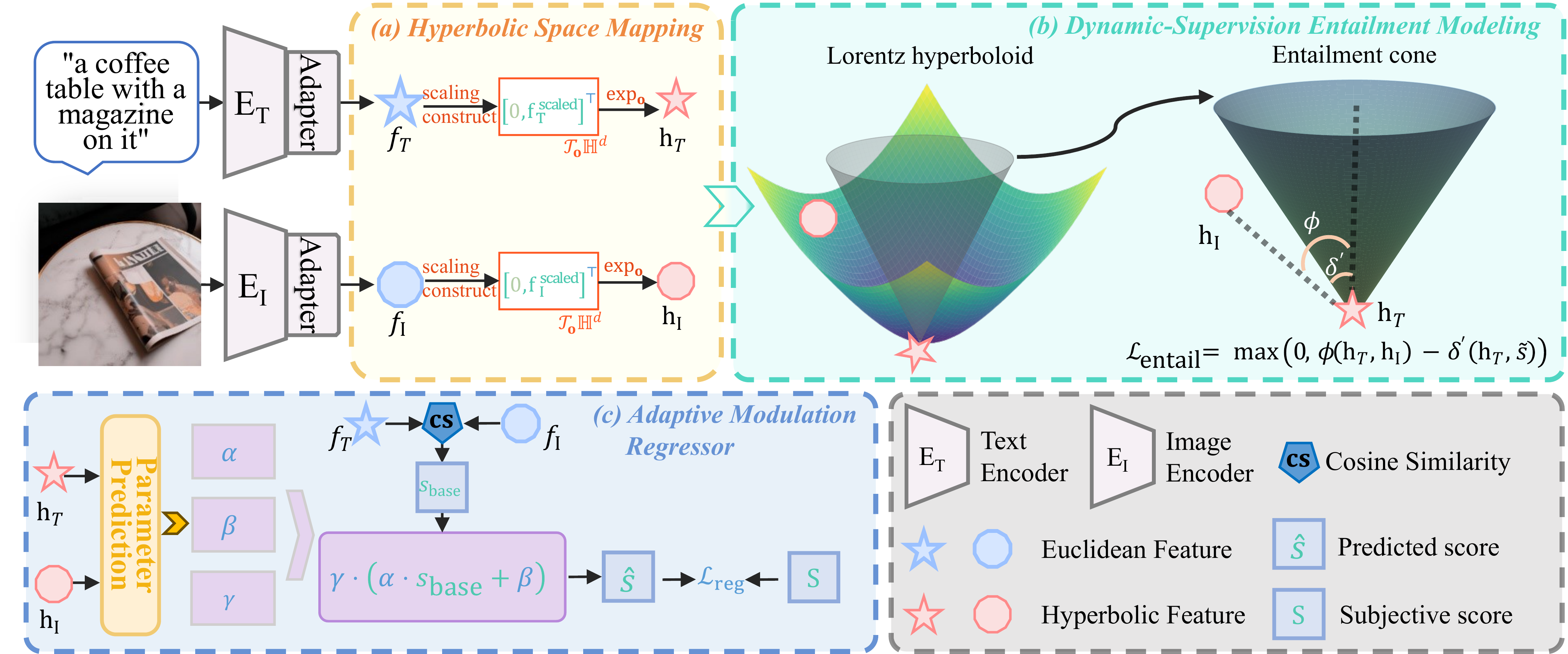}
\caption{Overall architecture of HyperAlign. CLIP with a lightweight Adapter extracts Euclidean features, which are then projected onto the Lorentz manifold to produce hyperbolic embeddings. During training, the normalized human score contracts the text entailment cone to impose geometric supervision; during inference, the regressor uses only score-free geometric primitives and Euclidean cosine similarity to predict the alignment score.}
\Description{The overall pipeline of HyperAlign showing CLIP feature extraction, hyperbolic space mapping, dynamic-supervision entailment modeling, and adaptive modulation regression.}
\label{fig:pipeline}
\end{figure*}

\section{Related Work}

\subsection{Text-to-Image Alignment Assessment Methods}

Traditional NS-IQA methods, although effective for measuring image distortions, fail to transfer effectively to AI-generated image scenarios. NS-IQA methods often assign high scores to pixel-level clear generated images, even if their content deviates from user-provided text descriptions. To bridge this perceptual gap, assessment methods for generated images have gradually emerged. Early assessment methods like IS~\cite{salimans2016is} and FID~\cite{heusel2017fid} measure overall distribution differences, but overlook image-text alignment. As generative paradigms evolved toward diffusion models, assessment focus has expanded beyond pixel-level distortions to incorporate semantic considerations, driving the emergence of AI-generated Image Quality Assessment (AG-IQA) methods. TIER~\cite{yuan2024tier} directly concatenates text and image features for quality score regression; IPCE~\cite{peng2024aigc} transforms assessment into weighted classification probability by designing text templates with different matching degrees.

However, these methods often fail to effectively decouple visual quality and image-text alignment, or incorrectly equate alignment simply with overall quality. In fact, these are two relatively independent and even potentially conflicting dimensions: a visually realistic image may be completely irrelevant to the text, and vice versa. Addressing this issue, recent research attempts to isolate alignment as an independent assessment dimension. IP-IQA~\cite{qu2024bringing} introduces Image2Prompt pre-training, using separate regression heads to predict quality and alignment respectively, achieving preliminary separation of assessment dimensions; Zhou et al. conducted in-depth exploration in this direction: they first proposed AMFF-Net~\cite{zhou2024adaptive}, utilizing adaptive feature fusion modules to integrate multi-scale image features, configuring three independent regression heads for visual quality, realism, and alignment; subsequently, they further proposed CIA-Net~\cite{zhou2025cia}, introducing cross-modal interactive attention modules that further enhanced joint prediction capability across these three dimensions through finer feature interaction. However, forcing simultaneous optimization of these two objectives within a shared-parameter encoder often leads to task interference. Furthermore, a greater limitation of these methods is the lack of explicit modeling of semantic hierarchical entailment relations, which results in limited generalization capability. Existing models primarily compute feature distances in Euclidean space, which essentially represents a flattened symmetric similarity metric. However, image-text correspondences are far more complex than simple similarity, with their core being directional asymmetric hierarchical entailment logic. This causes models to often ignore subtle image-text misalignments when facing complex fine-grained semantic combinations due to inability to distinguish hierarchical relations.

\subsection{Hyperbolic Vision-Language Models}

The negative curvature property of hyperbolic space enables low-distortion embedding of hierarchical structures. Compared to flat Euclidean space, hyperbolic space volume grows exponentially with radius. This advantage was first validated by Poincar\'{e} Embeddings~\cite{nickel2017poincare} on the WordNet~\cite{miller1995wordnet} hierarchy; subsequently, the Lorentz Model~\cite{nickel2018lorentz} improved numerical stability through Minkowski inner products.

Benefiting from excellent capabilities in representing hierarchical relations, hyperbolic geometry has been introduced to vision-language representation learning in recent years. MERU~\cite{desai2023meru} first elevated CLIP features to the Lorentz manifold, learning hierarchical representations by constraining text entails image partial order through entailment cones. HyCoCLIP~\cite{PalSDFGM2024} optimized compositional entailment for complex scenes; HyperVLM~\cite{srivastava2025hypervlm} applied hyperbolic modeling to multimodal understanding.
These works focus on pre-training and classification tasks, validating the effectiveness of hyperbolic space in modeling vision-language hierarchical relations. This paper explores its potential in fine-grained regression tasks by introducing a dynamic-supervision entailment modeling mechanism and adaptive modulation regressor, deeply integrating hyperbolic geometry with regression tasks, filling the gap in T2IAA, and effectively addressing the pain point that existing Euclidean space methods cannot precisely measure hierarchical semantic alignment.

\section{Method}

\subsection{Overall Framework}

Figure~\ref{fig:pipeline} summarizes HyperAlign. Given an image-text pair $(I,T)$, the model processes it in four steps. First, a pre-trained CLIP encoder together with a lightweight gated Adapter produces Euclidean embeddings $\mathbf{f}_I$ and $\mathbf{f}_T$. Detailed architecture of the gated Adapter is provided in the supplementary material. Second, a hyperbolic mapping module projects them onto the Lorentz manifold, yielding hyperbolic embeddings $\mathbf{h}_I$ and $\mathbf{h}_T$. Third, we compute three geometric primitives from $(\mathbf{h}_T,\mathbf{h}_I)$: hyperbolic distance, exterior angle, and text-cone aperture. During training only, the normalized human score $\tilde{s}$ contracts the text cone to provide dynamic geometric supervision. Finally, an adaptive modulation regressor uses the score-free geometric primitives to generate modulation parameters that guide Euclidean cosine similarity to regress to the final alignment score $\hat{s}$.

\subsection{Hyperbolic Space Mapping}

Hyperbolic space is well suited to hierarchical semantics because its volume grows exponentially with radius. We adopt the Lorentz model and use it to place more general concepts near the origin and more specific concepts farther away, which matches the asymmetric ``text contains image'' relation that Euclidean similarity cannot express explicitly.

\textbf{Lorentz Hyperbolic Manifold.} This paper adopts the Lorentz model to represent hyperbolic space. The Lorentz model $\mathbb{H}^{d}$ of $d$-dimensional hyperbolic space is defined as the upper sheet of a double-sheeted hyperboloid in $(d+1)$-dimensional Minkowski spacetime:

\begin{equation}
\mathbb{H}^{d} = \{\mathbf{x} \in \mathbb{R}^{d+1} : \langle \mathbf{x}, \mathbf{x} \rangle_\mathcal{L} = -\frac{1}{c}, c > 0\},
\end{equation}
where $c$ is the curvature parameter. For vector $\mathbf{x} = [x_0, x_1, \ldots, x_d]^{\top}$, following MERU's terminology, we call its first component $x_0$ the time component, and the last $d$ components $\mathbf{x}_{\text{space}} = [x_1, \ldots, x_d]^{\top} \in \mathbb{R}^d$ the space components, satisfying $x_0 = \sqrt{\frac{1}{c} + \|\mathbf{x}_{\text{space}}\|^2}$. $\langle \cdot, \cdot \rangle_\mathcal{L}$ is the Lorentzian inner product, defined as:
\begin{equation}
\langle \mathbf{x}, \mathbf{y} \rangle_\mathcal{L} = -x_0 y_0 + \sum_{i=1}^{d} x_i y_i.
\end{equation}

The distance in the Lorentz model is induced by the Lorentz inner product, defined as:

\begin{equation}
d_\mathcal{L}(\mathbf{x}, \mathbf{y}) = \frac{1}{\sqrt{c}} \cdot \text{arccosh}\left(-c \langle \mathbf{x}, \mathbf{y} \rangle_\mathcal{L}\right).
\end{equation}

The tangent space $\mathcal{T}_\mathbf{z}\mathbb{H}^{d}$ is a $d$-dimensional Euclidean linear approximation of the manifold at point $\mathbf{z}$, consisting of all vectors orthogonal to $\mathbf{z}$ under the Lorentz inner product: $\mathcal{T}_\mathbf{z}\mathbb{H}^{d} = \{\mathbf{v} \in \mathbb{R}^{d+1} : \langle \mathbf{z}, \mathbf{v} \rangle_\mathcal{L} = 0\}$.

\textbf{Exponential Map.} We lift Euclidean space features from tangent space to hyperbolic manifold through exponential mapping. Generally, for point $\mathbf{z} \in \mathbb{H}^{d}$ and vector $\mathbf{v} \in \mathcal{T}_\mathbf{z}\mathbb{H}^{d}$ in its tangent space, the exponential map $\exp_\mathbf{z}: \mathcal{T}_\mathbf{z}\mathbb{H}^{d} \to \mathbb{H}^{d}$ is defined as:
\begin{equation}
\exp_\mathbf{z}(\mathbf{v}) = \cosh\left(\sqrt{c} \|\mathbf{v}\|_\mathcal{L}\right) \mathbf{z} + \frac{\sinh\left(\sqrt{c} \|\mathbf{v}\|_\mathcal{L}\right)}{\sqrt{c} \|\mathbf{v}\|_\mathcal{L}} \mathbf{v},
\end{equation}
where $\|\mathbf{v}\|_\mathcal{L} = \sqrt{|\langle \mathbf{v}, \mathbf{v} \rangle_\mathcal{L}|}$ is the hyperbolic norm.

In our implementation, every Euclidean feature $\mathbf{f}$ is mapped in the same way. We first apply a learnable scaling factor for numerical stability:
\begin{equation}
\mathbf{f}_{\text{scaled}} = r \cdot \mathbf{f}, \quad r = \sigma(\omega) \cdot r_{\max},
\end{equation}
where $\sigma(\cdot)$ is the sigmoid function, $\omega$ is a learnable scalar, and $r_{\max}=1.0$ is the upper bound. Using the hyperbolic origin $\mathbf{o} = [\sqrt{1/c}, 0, \ldots, 0]^{\top}$ as the base point, we construct the tangent vector $\mathbf{u}=[0,\mathbf{f}_{\text{scaled}}]^{\top}\in\mathcal{T}_\mathbf{o}\mathbb{H}^{d}$ and obtain the hyperbolic embedding
\begin{equation}
\mathbf{h} = \exp_\mathbf{o}(\mathbf{u}).
\end{equation}
Applying this mapping to the image and text features produces $\mathbf{h}_I$ and $\mathbf{h}_T$, which are used by all subsequent geometric computations.

\subsection{Dynamic-Supervision Entailment Modeling}

Existing hyperbolic vision-language methods are mainly designed for binary or categorical decisions. These methods focus on whether entailment exists rather than capturing the degree of entailment, making them suitable for classification tasks but not directly applicable to continuous numerical regression tasks such as T2IAA. We would prefer highly aligned pairs to satisfy a stricter parent–child relation, while poorly aligned pairs should be penalized less strictly. To model this difference, we treat the text embedding as the parent concept, the image embedding as a candidate child concept, and during training we use human alignment score supervision to guide how strict the text cone should be.

\textbf{Hyperbolic Entailment Geometric Primitives.} As mentioned above, to enable the model to understand from a geometric perspective whether and to what extent the text semantics imply the presence of an image, we decouple image-text relations into interpretable geometric features. We describe the text-image relation using three interpretable quantities. The first is the text-cone half-aperture. For hyperbolic embeddings $\mathbf{h}_T$ and $\mathbf{h}_I$, the semantic range of text $\mathbf{h}_T$ is represented by an entailment cone with half-aperture angle $\delta(\mathbf{h}_T)$ that contracts as $\mathbf{h}_T$ moves away from the origin:
\begin{equation}
\delta(\mathbf{h}_T) = \arcsin \left( \frac{2k}{\sqrt{c} \|\mathbf{h}_{T,\text{space}}\|} \right),
\end{equation}
where $k$ is a constant controlling boundary conditions. This naturally encodes hierarchy: larger $\|\mathbf{h}_{T,\text{space}}\|$ yields narrower apertures that push aligned images outward, while general concepts near the origin maintain wide apertures accommodating diverse instances.
The second quantity is the exterior angle between text and image, which measures whether the image falls inside the cone. We compute exterior angle $\phi(\mathbf{h}_T, \mathbf{h}_I)$ at vertex $\mathbf{h}_T$ in triangle $\Delta O\mathbf{h}_T\mathbf{h}_I$. By hyperbolic law of cosines with $\eta = \langle \mathbf{h}_T, \mathbf{h}_I \rangle_{\mathcal{L}}$:
\begin{equation}
\phi(\mathbf{h}_T, \mathbf{h}_I) = \arccos\left( \frac{h_{I,0} + h_{T,0} c \eta}{ \|\mathbf{h}_{T,\text{space}}\| \sqrt{(c\eta)^2 - 1} }\right),
\end{equation}
where $h_{T,0}, h_{I,0}$ are time components, $\mathbf{h}_{T,\text{space}}$ are space components, and $\langle \cdot, \cdot \rangle_{\mathcal{L}}$ is the Lorentzian inner product. Smaller $\phi$ indicates stronger semantic alignment.
The third quantity is the pairwise hyperbolic distance $d_\mathcal{L}(\mathbf{h}_T, \mathbf{h}_I)$, which complements the angle-based relation with a global measure of separation.

\begin{figure}[t]
\centering
\includegraphics[width=0.5\textwidth]{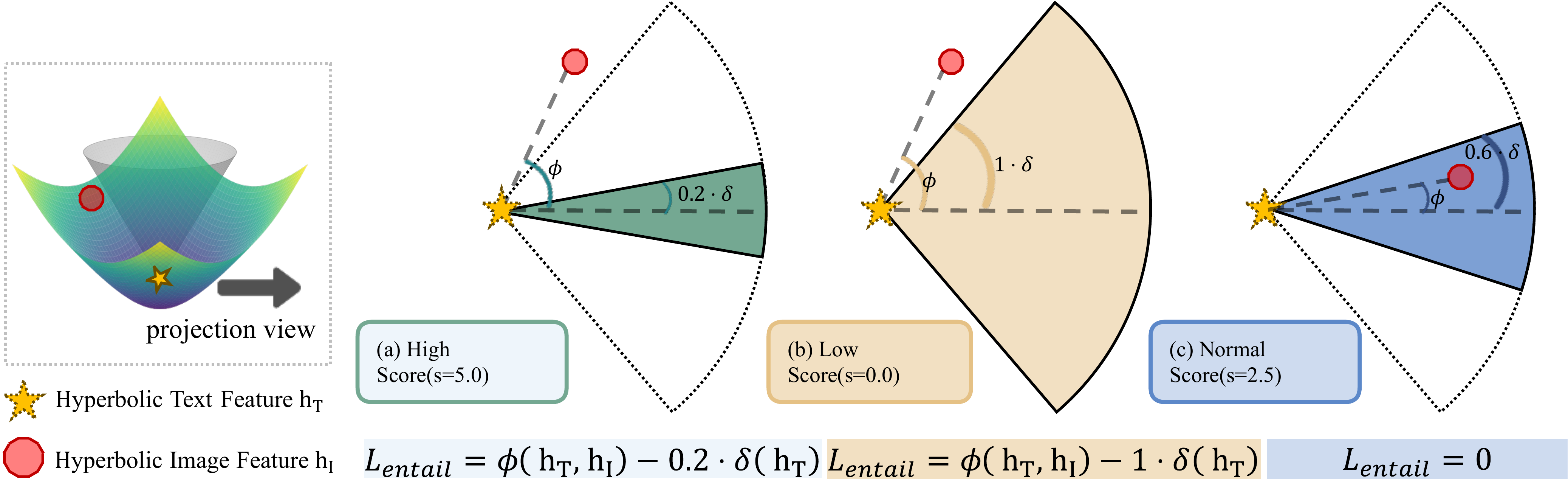}
\caption{Visualization of dynamic-supervision entailment cones. High-alignment samples are constrained within narrow cones, while low-alignment samples are allowed wider cones.}
\Description{Diagram showing entailment cones with dynamic apertures controlled by alignment scores: narrow cones for high-alignment and wide cones for low-alignment samples.}
\label{fig:entail_loss}
\end{figure}

\textbf{Training-Time Dynamic Supervision.} As illustrated in Figure~\ref{fig:entail_loss}, we do not use a fixed cone for every sample. Instead, during training we use the normalized ground-truth score $\tilde{s}\in[0,1]$ to contract the text cone:
\begin{equation}
\psi(\tilde{s}) = 1 - \kappa \cdot \tilde{s}, \quad \delta_{\text{dyn}}(\mathbf{h}_T, \tilde{s}) = \psi(\tilde{s}) \cdot \delta(\mathbf{h}_T),
\end{equation}
where $\kappa$ is the cone-contraction coefficient. Larger alignment scores yield smaller $\psi(\tilde{s})$, which produces a narrower cone and therefore imposes a stricter entailment constraint. The resulting loss is
\begin{equation}
\mathcal{L}_{\text{entail}}(\mathbf{h}_T, \mathbf{h}_I, \tilde{s}) = \max\left(0, \phi(\mathbf{h}_T, \mathbf{h}_I) - \delta_{\text{dyn}}(\mathbf{h}_T, \tilde{s})\right).
\end{equation}
Minimizing this loss pushes high-alignment samples into tighter cones while leaving more tolerance for low-alignment samples. In this way, discrete entailment geometry becomes a continuous supervision signal that reflects score differences. Importantly, $\tilde{s}$ appears only in this training-time branch where the loss needs to be calculated, it is not needed when the model is deployed for prediction.

\subsection{Adaptive Modulation Regressor}
Although hyperbolic space can effectively characterize hierarchical entailment relations between text and images, it is not ideal as the sole regression space. The exponential mapping required to project Euclidean representations into hyperbolic space may weaken some semantic priors accumulated by pre-trained models, and hyperbolic metrics are highly sensitive to numerical variations, making them unsuitable for direct score regression. To address this limitation, we propose an adaptive modulation regressor that retains Euclidean cosine similarity as the base metric while leveraging hyperbolic geometric features to modulate it dynamically. This design avoids the heavy multi-scale or cross-attention operations adopted by some competing methods, requiring only a compact network to generate modulation parameters from the three geometric primitives. Specifically, we first compute the base similarity score:
\begin{equation}
s_{\text{base}} = \mathrm{CosineSimilarity}(\mathbf{f}_I, \mathbf{f}_T).
\end{equation}
We then form a score-free geometric descriptor
\begin{equation}
\mathcal{Z} = \left\{ d_\mathcal{L}(\mathbf{h}_I,\mathbf{h}_T), \phi(\mathbf{h}_T,\mathbf{h}_I), \delta(\mathbf{h}_T) \right\},
\end{equation}
and feed it into a lightweight parameter prediction network $\mathcal{M}(\cdot)$. The logic flow is shown in Algorithm~\ref{alg:regression}.
\begin{algorithm}[tb]
\small
\caption{Adaptive Modulation Regressor}
\label{alg:regression}
\begin{algorithmic}[1]
\REQUIRE Euclidean features $\mathbf{f}_I, \mathbf{f}_T$;
         Hyperbolic primitives $\mathcal{Z} = \{d_\mathcal{L}, \phi, \delta\}$
\ENSURE Predicted alignment score $\hat{s}$
\STATE $s_{\text{base}} \leftarrow \text{CosineSimilarity}(\mathbf{f}_I, \mathbf{f}_T)$
\STATE $\alpha, \beta, \gamma \leftarrow \mathcal{M}(\mathcal{Z})$
\STATE $\hat{s} \leftarrow \gamma \cdot (\alpha \cdot s_{\text{base}} + \beta)$
\RETURN $\hat{s}$
\end{algorithmic}
\end{algorithm}
The three outputs of $\mathcal{M}$ have clear roles. Scale $\alpha$ adjusts the dynamic range of cosine similarity, allowing highly aligned samples to be separated more clearly. Bias $\beta$ compensates for systematic offsets, for example when the Euclidean similarity is reasonable but the hyperbolic relation indicates a boundary case. Confidence $\gamma$ acts as a gate that down-weights samples with uncertain or contradictory geometry. Because $\mathcal{Z}$ does not depend on any ground-truth score, this regressor can be used directly at inference time.

\subsection{Training and Inference Workflow}

HyperAlign is trained by combining prediction accuracy with geometric supervision. The total loss is
\begin{equation}
  \mathcal{L}_{\text{total}} = \mathcal{L}_{\text{reg}} + \lambda \cdot \mathcal{L}_{\text{entail}},
\end{equation}
where $\lambda$ balances the two objectives and the regression loss is
\begin{equation}
\mathcal{L}_{\text{reg}} = \|\hat{s} - s\|_1.
\end{equation}
During training, each sample goes through both branches: the prediction branch outputs $\hat{s}$, and the supervision branch uses the normalized score $\tilde{s}$ to contract the text cone and compute $\mathcal{L}_{\text{entail}}$. During inference, the supervision branch is removed. The model only computes $\mathbf{f}_I,\mathbf{f}_T$, maps them to $\mathbf{h}_I,\mathbf{h}_T$, forms $\mathcal{Z}$, and predicts $\hat{s}$. Therefore, HyperAlign does not require any ground-truth score or score-conditioned cone at test stage.

\begin{table*}[t]
  \centering
  \small
  \caption{Single-database performance comparison. Best results are in \textbf{bold}, second-best are \underline{underlined}.}
  \label{tab:in_database}
  \begin{tabular*}{\textwidth}{@{\extracolsep{\fill}}l c ccccccc}
    \toprule
    \multirow{2}{*}{Method} & \multirow{2}{*}{Venue} & \multicolumn{2}{c}{AGIQA-3K} & \multicolumn{2}{c}{AIGCIQA2023} & \multicolumn{2}{c}{PKU-I2IQA} & \multirow{2}{*}{Avg.} \\
    \cmidrule(lr){3-4} \cmidrule(lr){5-6} \cmidrule(lr){7-8}
     & & SRCC & PLCC & SRCC & PLCC & SRCC & PLCC &  \\
    \midrule
    DB-CNN~\cite{zhang2020blind} & TCSVT'20 & 0.6329 & 0.7823 & 0.6837 & 0.6787 & 0.6083 & 0.5925 & 0.6631 \\
    HyperIQA~\cite{su2020blindly} & CVPR'20 & 0.6276 & 0.8087 & 0.7541 & 0.7439 & 0.7239 & 0.7062 & 0.7274 \\
    MUSIQ~\cite{ke2021musiq} & ICCV'21 & 0.6292 & 0.7839 & 0.7620 & 0.7527 & 0.6379 & 0.6531 & 0.7031 \\
    TReS~\cite{golestaneh2022no} & TIP'22 & 0.6366 & 0.8134 & 0.7292 & 0.7266 & 0.6480 & 0.6456 & 0.6999 \\
    Re-IQA~\cite{saha2023re} & CVPR'23 & 0.6373 & 0.7880 & 0.6430 & 0.6355 & 0.5705 & 0.5690 & 0.6406 \\
    StairIQA~\cite{sun2023blind} & JSTSP'23 & 0.6348 & 0.8006 & 0.6641 & 0.6625 & 0.5739 & 0.5720 & 0.6513 \\
    LIQE~\cite{zhang2023liqe} & CVPR'23 & 0.6516 & 0.7404 & 0.7705 & 0.7589 & 0.7627 & 0.7115 & 0.7326 \\
    CLIPIQA~\cite{wang2023exploring} & AAAI'23 & 0.6456 & 0.6607 & 0.7321 & 0.6826 & 0.7071 & 0.6148 & 0.6738 \\
    \cdashline{1-9}
    \noalign{\vskip 3pt}
    IP-IQA~\cite{qu2024bringing} & ICME'24 & 0.7578 & 0.8544 & - & - & - & - & - \\
    AMFF-Net~\cite{zhou2024adaptive} & TBC'24 & 0.7513 & 0.8476 & 0.7782 & 0.7638 & 0.7796 & 0.7708 & 0.7819 \\
    IPCE~\cite{peng2024aigc} & CVPRW'24 & 0.7697 & \underline{0.8725} & \underline{0.7979} & \underline{0.7887} & - & - & - \\
    CIA-Net~\cite{zhou2025cia} & PR'25 & \underline{0.7797} & 0.8687 & 0.7977 & 0.7800 & \underline{0.7855} & \underline{0.7829} & \underline{0.7991} \\
    \midrule
    HyperAlign (ours) & - & \textbf{0.7927} & \textbf{0.8830} & \textbf{0.8078} & \textbf{0.8013} & \textbf{0.7977} & \textbf{0.8316} & \textbf{0.8190} \\
    \bottomrule
  \end{tabular*}
\end{table*}

\section{Experiments}

\subsection{Experimental Setup}

\textbf{Datasets.} We conduct evaluations on three mainstream benchmark datasets. \textbf{AGIQA-3K}~\cite{li2023agiqa} comprises 2982 AI-generated images from six generative models, covering diverse prompt types including objects, scenes, and styles, with human-annotated alignment Mean Opinion Scores (MOS) ranging from 1 to 5. \textbf{AIGCIQA2023}~\cite{kou2023aigciqa} consists of 2400 images generated by eight mainstream models, with prompts spanning multiple topic categories and alignment MOS ranging from 1 to 5. \textbf{PKU-I2IQA}~\cite{zhou2023i2iqa} contains 2000 image pairs with alignment MOS ranging from 1 to 5. For single-database evaluation, following previous works, AGIQA-3K is randomly divided into training and testing subsets with a ratio of 4:1, ensuring identical text prompts appear only within the same subset to prevent content overlap. For AIGCIQA2023 and PKU-I2IQA, we use 3:1 train-test splits. We report average results over ten random splits.

\textbf{Evaluation Metrics.} We use two widely adopted metrics: Spearman Rank Correlation Coefficient (SRCC) and Pearson Linear Correlation Coefficient (PLCC). The SRCC measures the monotonic relationship between predicted and ground-truth scores, robust to outliers. The PLCC measures the linear goodness-of-fit between predicted and true values, more sensitive to absolute prediction accuracy. Both range from -1 to 1, with values closer to 1 indicating better model performance.

\textbf{Implementation Details.} We use ViT-B/16 CLIP with batch size 8 and AdamW optimizer, adopting grouped learning rates: CLIP backbone $2 \times 10^{-6}$ with weight decay $1 \times 10^{-4}$, Adapter and regressor $4 \times 10^{-4}$ with weight decay 0.005. Learning rate scheduling uses StepLR strategy, decaying to 0.5 of the original every 10 epochs. L1 loss with entailment loss weight $\lambda=0.2$. Training employs mixed-precision to accelerate computation, with maximum epochs set to 20 and early stopping strategy that terminates training when validation set SRCC does not improve for 6 consecutive epochs. All experiments are conducted on a single NVIDIA GeForce GTX4090 GPU.

\begin{figure}[t]
\centering
\includegraphics[width=\linewidth]{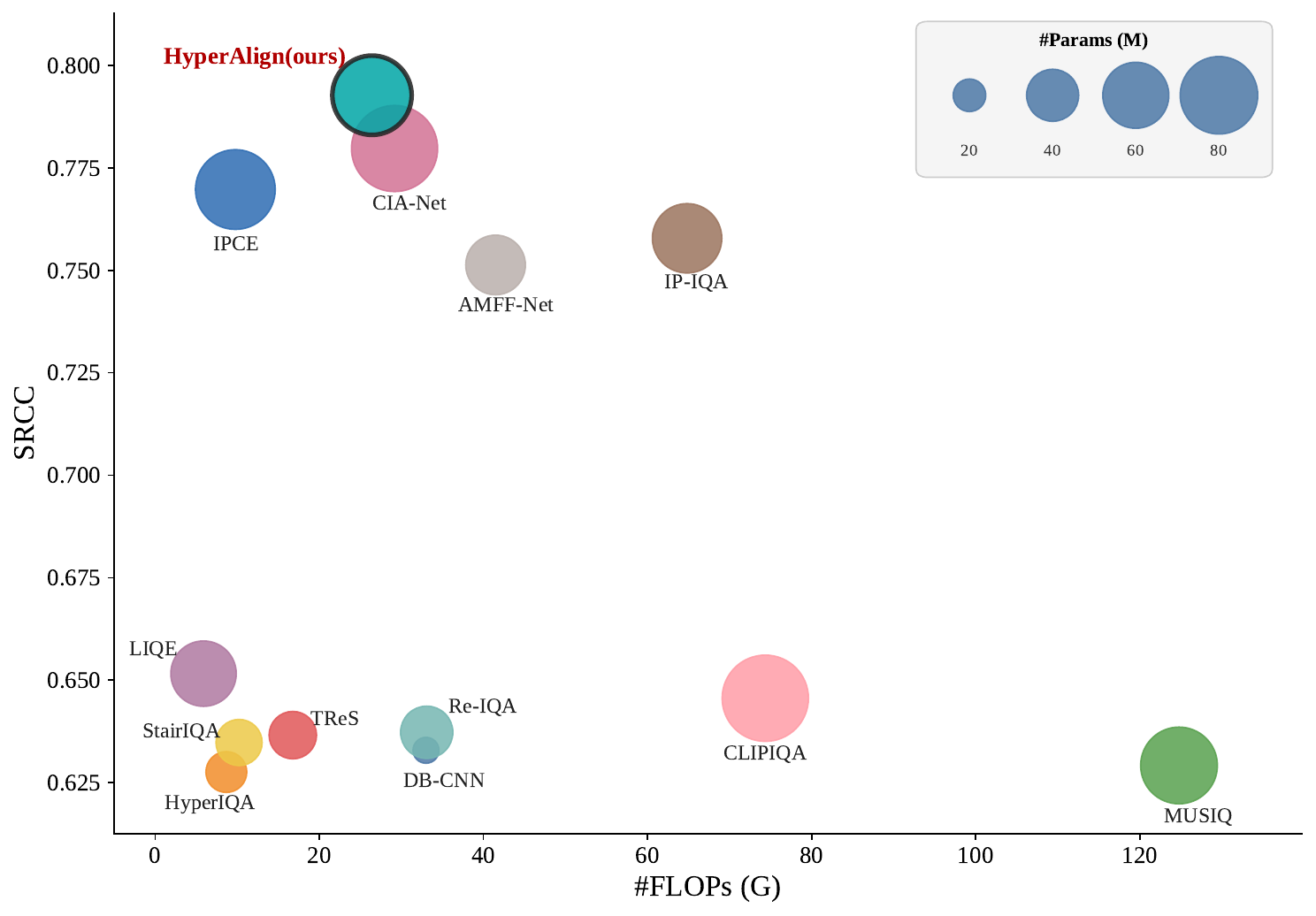}
\caption{Computational cost versus prediction accuracy. The horizontal axis represents FLOPs, the vertical axis represents SRCC on AGIQA-3K, and the bubble size indicates the number of parameters.}
\Description{Bubble chart comparing FLOPs, SRCC, and parameter counts across different methods, showing that HyperAlign achieves the highest SRCC with moderate computational cost.}
\label{fig:bubble_performance}
\end{figure}

\subsection{Single-database Performance Evaluation}
Following the experimental setup described above, we validate HyperAlign's effectiveness on T2IAA by comparing against 12 mainstream methods, including 8 NS-IQA and 4 AG-IQA methods. As shown in Table~\ref{tab:in_database}, the results reveal a substantial performance gap between traditional NS-IQA and AG-IQA methods, confirming that single-modality methods face significant challenges when directly applied to generative assessment. In contrast, HyperAlign achieves state-of-the-art performance across all datasets with an average SRCC of 0.8190, surpassing the second-best method CIA-Net by 1.99 percentage points. HyperAlign's superior performance is fundamentally attributed to the synergistic effect of hyperbolic entailment modeling and adaptive modulation regression. Specifically, hyperbolic space's hierarchical modeling capability enables HyperAlign to capture asymmetric entailment relations across different semantic levels, explicitly encoding whether generated images satisfy the hierarchical containment logic required by text prompts. Building upon this geometric foundation, the adaptive modulation regressor further leverages entailment geometric features to generate sample-level calibration parameters, enabling personalized score mapping that effectively captures fine-grained alignment characteristics between diverse image content and text prompts. By tightly coupling geometric structure constraints with adaptive score calibration, HyperAlign achieves significant performance gains across all benchmarks.

Although HyperAlign introduces several components on top of CLIP, including Hyperbolic Space Mapping and an Adaptive Modulation Regressor, these components are intentionally designed to be lightweight. Figure~\ref{fig:bubble_performance} presents the computational complexity of different methods. HyperAlign requires 26.42G FLOPs and contains 83.25M parameters, where the FLOPs and parameter counts are measured using the THOP library. Compared with most competing methods, HyperAlign ranks among the most efficient in terms of FLOPs, requiring only about one-fifth of MUSIQ and approximately one-third of CLIPIQA and IP-IQA.
In terms of parameter count, although it is higher than some traditional IQA methods, HyperAlign remains more lightweight than several CLIP-based methods, such as CIA-Net, IPCE, and CLIPIQA. This efficiency stems from the architectural design of HyperAlign: the hyperbolic mapping and entailment geometry computations involve only lightweight algebraic operations, while the adaptive modulation regressor generates modulation parameters through a compact network, avoiding the heavy multi-scale or cross-attention operations adopted by some competing methods.
Despite its efficiency, HyperAlign achieves superior prediction accuracy compared with all competing methods, including CIA-Net with 97.09M parameters and AMFF-Net with 41.48G FLOPs, further demonstrating the effectiveness of our approach. Overall, HyperAlign achieves a favorable balance between performance and efficiency.

\begin{figure}[t]
    \centering
    \includegraphics[width=\linewidth]{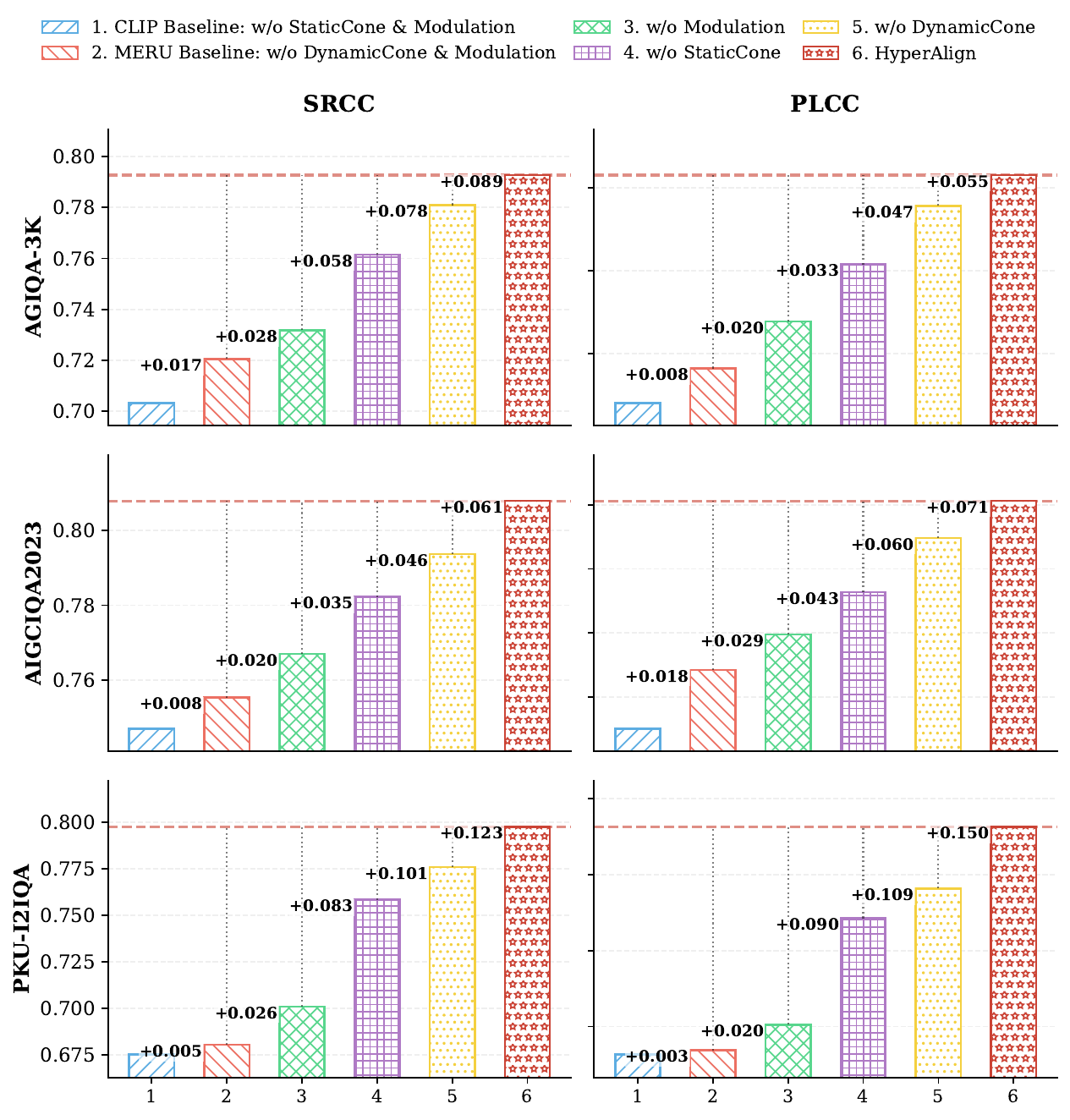}

    \caption{
        Ablation study of HyperAlign on three standard datasets.
    }
    \Description{Bar charts showing ablation study results across six configurations on three datasets, demonstrating progressive performance improvements.}
    \label{fig:ablation}
\end{figure}

\subsection{Ablation Studies}
To analyze the contribution of each component in HyperAlign more thoroughly, we conduct a series of ablation experiments. We use the same configuration as in the single-database evaluation, repeat each experiment 10 times, and report the average results. Specifically, we investigate the following three key components: 1) static entailment cone modeling without alignment-score guidance (StaticCone), 2) dynamic entailment cone modeling with alignment-score guidance (DynamicCone), and 3) the adaptive modulation regressor (Modulation).

Figure~\ref{fig:ablation} reveals a progressive performance trend across the six configurations. Configuration (1) can be regarded as the CLIP baseline: it contains neither entailment cone modeling nor the modulation regressor, and predicts the alignment score through a simple mapping. Configuration (2), equipped with a static entailment cone, can be regarded as the MERU baseline, as it includes only static entailment cone modeling without the modulation regressor. Under this setting, SRCC improves over baseline (1) by +0.017/+0.008/+0.005 on AGIQA-3K/AIGCIQA2023/PKU-I2IQA, confirming that hyperbolic entailment geometry itself already provides a useful hierarchical inductive bias. Replacing the static cone in configuration (2) with the dynamic-supervision cone in configuration (3) further increases the gains to +0.028/+0.020/+0.026, demonstrating that the adaptive aperture of our proposed dynamic cone is better suited to the alignment assessment task than the fixed geometric structure of the static cone adopted in MERU. Configuration (4), which applies modulation without any entailment cone, achieves substantial improvements of +0.058, +0.035, and +0.083 across the three datasets. This may be attributed to two factors: the latent hierarchical semantics already present in CLIP features after hyperbolic projection, and the positive effect of sample-level adaptive parameters that enable personalized score mapping.
In configurations (5) and (6), the synergistic effect becomes evident: reintroducing entailment cones on top of modulation yields substantial additional gains. Configuration (5) with the static cone reaches +0.078/+0.046/+0.101, while the full HyperAlign model (6) with the dynamic cone achieves peak improvements of +0.089/+0.061/+0.123. This indicates that the improvement of HyperAlign cannot be attributed solely to hyperbolic entailment cone modeling. More importantly, it stems from our novel integration of dynamic supervision and adaptive modulation, which specifically enhances the effectiveness of hyperbolic modeling for this regression task. Finally, an interesting observation is that the dynamic cone consistently outperforms the static cone both without modulation (configuration 3 vs. 2: +0.011/+0.012/+0.021) and with modulation (configuration 6 vs. 5: +0.012/+0.014/+0.022), further confirming its superiority. These results demonstrate that explicit geometric supervision and feature-guided modulation complement each other and jointly establish an optimal hierarchical structure for alignment assessment.

\begin{table}[t]
  \centering
  \small
  \caption{Sensitivity analysis of entailment loss weight $\lambda$ across all datasets.}
  \label{tab:lambda_sensitivity}
  \begin{tabular*}{\linewidth}{@{\extracolsep{\fill}}c cccccc c}
    \toprule
    \multirow{2}{*}{$\lambda$} & \multicolumn{2}{c}{AGIQA-3K} & \multicolumn{2}{c}{AIGCIQA2023} & \multicolumn{2}{c}{PKU-I2IQA} & \multirow{2}{*}{Avg.} \\
    \cmidrule(lr){2-3} \cmidrule(lr){4-5} \cmidrule(lr){6-7}
     & SRCC & PLCC & SRCC & PLCC & SRCC & PLCC & \\
    \midrule
    0.0 & 0.7615 & 0.8615 & 0.8021 & 0.7936 & 0.7785 & 0.7904 & 0.7979 \\
    \textbf{0.2} & \textbf{0.7927} & \textbf{0.8830} & 0.8078 & 0.8013 & 0.7977 & 0.8316 & \textbf{0.8190} \\
    0.4 & 0.7884 & 0.8811 & 0.8075 & 0.8000 & 0.7990 & 0.8326 & 0.8181 \\
    0.6 & 0.7897 & 0.8815 & 0.8019 & 0.7946 & 0.7997 & \textbf{0.8338} & 0.8169 \\
    0.8 & 0.7856 & 0.8796 & \textbf{0.8117} & \textbf{0.8042} & 0.7913 & 0.8246 & 0.8162 \\
    1.0 & 0.7835 & 0.8784 & 0.8114 & 0.7949 & \textbf{0.8003} & 0.8253 & 0.8156 \\
    \bottomrule
  \end{tabular*}
\end{table}

\begin{table}[t]
  \centering
  \small
  \caption{Sensitivity analysis of cone-contraction coefficient $\kappa$ across all datasets.}
  \label{tab:kappa_sensitivity}
  \begin{tabular*}{\linewidth}{@{\extracolsep{\fill}}c cccccc c}
    \toprule
    \multirow{2}{*}{$\kappa$} & \multicolumn{2}{c}{AGIQA-3K} & \multicolumn{2}{c}{AIGCIQA2023} & \multicolumn{2}{c}{PKU-I2IQA} & \multirow{2}{*}{Avg.} \\
    \cmidrule(lr){2-3} \cmidrule(lr){4-5} \cmidrule(lr){6-7}
     & SRCC & PLCC & SRCC & PLCC & SRCC & PLCC & \\
    \midrule
    0.0 & 0.7809 & 0.8756 & \textbf{0.8087} & 0.8012 & 0.7971 & 0.8307 & 0.8157 \\
    0.2 & 0.7879 & 0.8791 & 0.8070 & 0.7999 & 0.7991 & \textbf{0.8327} & 0.8176 \\
    0.4 & 0.7904 & \textbf{0.8837} & 0.8079 & 0.8001 & 0.7906 & 0.8325 & 0.8175 \\
    0.6 & 0.7886 & 0.8799 & 0.8077 & 0.8005 & \textbf{0.7997} & 0.8326 & 0.8182 \\
    \textbf{0.8} & \textbf{0.7927} & 0.8830 & 0.8078 & \textbf{0.8013} & 0.7977 & 0.8316 & \textbf{0.8190} \\
    1.0 & 0.7868 & 0.8804 & 0.8027 & 0.8002 & 0.7973 & 0.8293 & 0.8161 \\
    \bottomrule
  \end{tabular*}
\end{table}

\subsection{Hyperparameter Sensitivity Analysis}
Beyond standard training configurations, HyperAlign introduces two method-specific hyperparameters: the entailment loss weight $\lambda$ and the cone-contraction coefficient $\kappa$ in the dynamic modulation factor $\psi(\tilde{s}) = 1 - \kappa \cdot \tilde{s}$. To verify the robustness of our method, we vary each independently while fixing the other at its optimum. As shown in Table~\ref{tab:lambda_sensitivity}, setting $\lambda=0$ drops the average from 0.8190 to 0.7979, confirming that geometric constraints are indispensable. Performance peaks at $\lambda=0.2$ and degrades gracefully beyond it, with even $\lambda=1.0$ still reaching 0.8156, indicating that the entailment loss complements rather than conflicts with the regression objective. For $\kappa$ in Table~\ref{tab:kappa_sensitivity}, setting $\kappa=0$ disables dynamic cone adjustment and reduces the model to the static-cone variant, yielding 0.8157. Increasing $\kappa$ progressively tightens cones for high-alignment samples; performance peaks at $\kappa=0.8$ (0.8190) and slightly drops at $\kappa=1.0$ (0.8161), because the contraction factor approaches zero for $\tilde{s}\approx 1$ and nearly collapses the cone. The results confirm that HyperAlign's gains stem from its architectural design rather than sensitive hyperparameter selection, and we adopt $\lambda=0.2$ and $\kappa=0.8$ as defaults.

\begin{figure*}[!htbp]
\centering
\includegraphics[width=0.9\textwidth]{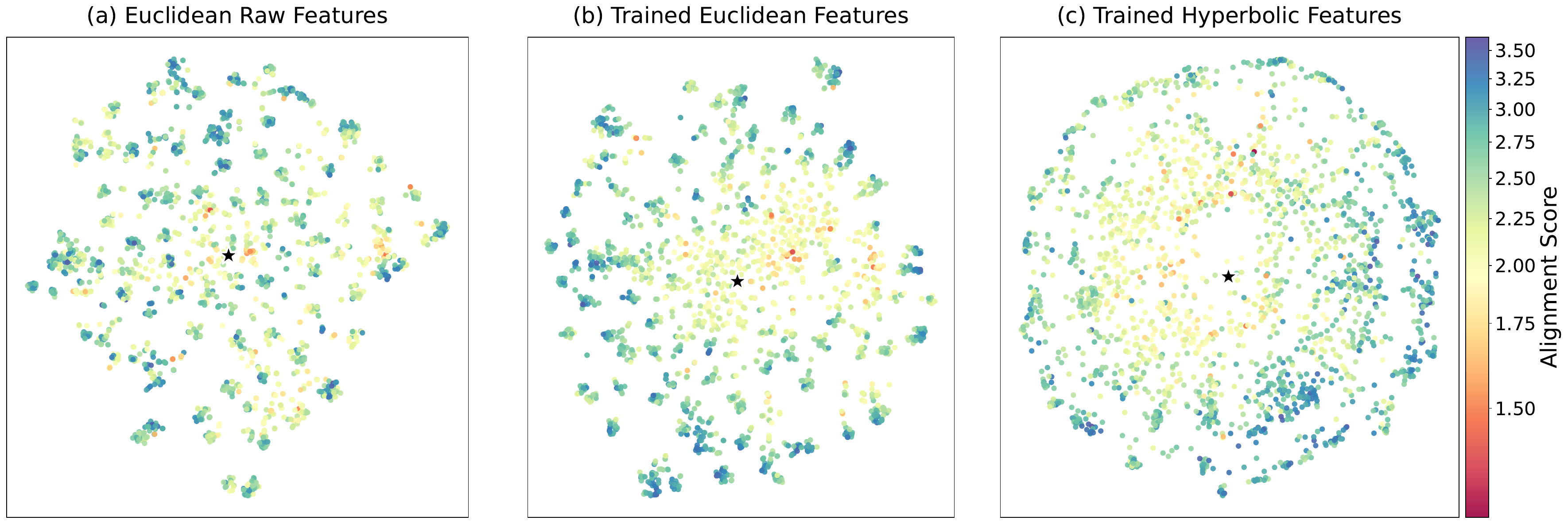}
\caption{
Feature space visualization on AIGCIQA2023 dataset using t-SNE for Euclidean features and CO-SNE for hyperbolic features.
}
\Description{Three t-SNE and CO-SNE visualizations comparing CLIP original features, trained Euclidean features, and trained hyperbolic features, showing progressive improvement in hierarchical separation.}
\label{fig:visual_exp}
\end{figure*}

\subsection{Visualization Analysis}

To intuitively demonstrate the improvement effect of hyperbolic space mapping on feature representation, we conduct image feature space visualization experiments on the AIGCIQA2023 dataset. We focus on visualizing image features because they directly reflect how the model organizes visual concepts according to their alignment degrees with prompts, providing clearer insights into the learned hierarchical structure. Specifically, we compare three types of image feature representations: 1) CLIP original features with directly loaded pre-trained weights without downstream task training, 2) trained Euclidean features fine-tuned on the downstream alignment assessment task, and 3) trained hyperbolic features fine-tuned on the downstream alignment assessment task. For Euclidean features, we apply t-SNE for dimensionality reduction; for hyperbolic features, we apply CO-SNE~\cite{guo2022cosne}, which preserves distances and more accurately maintains hyperbolic geometric structure.

As shown in Figure~\ref{fig:visual_exp}, CLIP original features demonstrate certain clustering capability but lack hierarchical stratification across different score levels. Trained Euclidean features show improved separation with emerging hierarchical tendencies, yet the stratification remains subtle with considerable overlap among score groups. In contrast, trained hyperbolic features reveal pronounced radial hierarchical structure: low-score samples in warm colors concentrate near the origin, while high-score samples in cool colors disperse toward the disk boundary, forming distinct hierarchical layers. This aligns with hyperbolic geometric properties where distance from the origin encodes semantic specificity---nodes farther from the origin with larger norms correspond to leaf-level specific concepts, while those near the origin represent abstract concepts. This visualization geometrically validates the effectiveness of entailment loss and feature guidance in HyperAlign, demonstrating that hyperbolic space naturally provides an ideal geometric foundation for hierarchical alignment modeling. Additional visualizations on AGIQA-3K and PKU-I2IQA are provided in the supplementary material and show consistent hierarchical patterns.

\begin{table}[t]
  \centering
  \small
  \setlength{\tabcolsep}{1pt}
  \caption{Comparison of results on cross-database testing. Best results are in \textbf{bold}, second-best are \underline{underlined}. Avg. denotes the mean of SRCC and PLCC over both transfer directions.}
  \label{tab:cross_database}
  \begin{tabular*}{\linewidth}{@{\extracolsep{\fill}}lccccc}
    \toprule
    \multirow{3}{*}{Method} & \multicolumn{2}{c}{\makecell{AIGCIQA2023\\ $\to$ AGIQA3K}} & \multicolumn{2}{c}{\makecell{AGIQA3K$\to$ \\ AIGCIQA2023}} & \multirow{3}{*}{Avg.} \\
    \cmidrule(lr){2-3} \cmidrule(lr){4-5}
     & SRCC & PLCC & SRCC & PLCC &  \\
    \midrule
    DB-CNN & 0.3900 & 0.4350 & 0.4700 & 0.4600 & 0.4388 \\
    HyperIQA & 0.4180 & 0.4650 & 0.4640 & 0.4310 & 0.4445 \\
    MUSIQ & 0.3940 & 0.4370 & 0.5250 & 0.5150 & 0.4678 \\
    TRES & 0.4450 & 0.4880 & 0.5050 & 0.4830 & 0.4803 \\
    Re-IQA & 0.2430 & 0.1540 & 0.4790 & 0.4840 & 0.3400 \\
    LIQE & 0.5002 & 0.4830 & 0.4927 & 0.5188 & 0.4987 \\
    CLIPIQA & 0.4809 & 0.4409 & 0.4300 & 0.4400 & 0.4480 \\
    AMFF-Net & \underline{0.5537} & \underline{0.6240} & 0.5461 & 0.5485 & 0.5681 \\
    CIA-Net & 0.5287 & 0.5498 & \textbf{0.6506} & \textbf{0.7443} & \underline{0.6184} \\
    \textbf{Ours} & \textbf{0.7013} & \textbf{0.7926} & \underline{0.6309} & \underline{0.6244} & \textbf{0.6873} \\
    \bottomrule
  \end{tabular*}
\end{table}

\subsection{Cross-Database Performance Evaluation}
To evaluate HyperAlign's generalization capability, we conduct cross-database experiments by training on one dataset and testing on another. Specifically, we examine two scenarios: 1) training on AIGCIQA2023 and testing on AGIQA-3K; 2) training on AGIQA-3K and testing on AIGCIQA2023. PKU-I2IQA is excluded as it involves image to image rather than only text to image. Each experiment is repeated with different random seeds and we report the average results. Table~\ref{tab:cross_database} shows that cross-database evaluation is substantially harder than single-database evaluation for all methods, confirming the severity of domain shift. Nevertheless, HyperAlign attains the best overall average correlation across both directions and both metrics, reaching \textbf{0.6873} and outperforming CIA-Net (0.6184) by 6.89 percentage points. We attribute this advantage to the fact that HyperAlign does not rely solely on dataset-specific similarity statistics in Euclidean space. Instead, it explicitly models the hierarchical and asymmetric entailment structure between text and image in hyperbolic space, which provides a more transferable criterion for judging whether image content truly satisfies prompt semantics under distribution shift. As a result, the model is better able to preserve semantically meaningful alignment cues across datasets, leading to stronger overall cross-database robustness.

\section{Conclusion}

This paper has proposed HyperAlign, an image-text alignment assessment framework based on hyperbolic space, utilizing hyperbolic geometry to model the hierarchical entailment relations between images and text. By designing a dynamic-supervision entailment modeling mechanism in hyperbolic space, we have effectively captured the hierarchical semantic structures inherent in image-text pairs. Ablation experiments have validated the synergistic effect of entailment loss and entailment feature guidance, indicating that both components complementarily promote performance improvement. Extensive experiments on three AIGC assessment benchmarks have demonstrated that HyperAlign outperforms existing methods while maintaining low computational cost. In the future, we will explore integrating large language models into our method to develop alignment assessment methods with more interpretable outputs.


\bibliographystyle{ACM-Reference-Format}
\bibliography{references}

\end{document}